\documentclass[sigconf,natbib=false]{acmart}

\AtBeginDocument{%
  }

\setcopyright{acmlicensed}
\copyrightyear{2024}
\acmYear{2024}
\acmDOI{XXXXXXX.XXXXXXX}

\acmConference{CONSEQUENCES Workshop at ACM RecSys (CONSEQUENCES@RecSys’24)} {October 14, 2024}{Bari, Italy}

\usepackage{algorithm}
\usepackage{algpseudocode}

\begin{document}

\title[Adaptive Mixture Importance Sampling]{Adaptive Mixture Importance Sampling for Automated Ads Auction Tuning}


\author{Yimeng Jia}
\orcid{0009-0007-1265-1555}
\affiliation{%
  \country{yimengjia@microsoft.com}
}

\author{Kaushal Paneri}
\orcid{0000-0002-8785-0723}
\affiliation{%
  \country{kapaneri@microsoft.com}
}

\author{Rong Huang}
\affiliation{%
  \country{huangr@microsoft.com}
}

\author{Kailash Singh Maurya}
\affiliation{%
  \country{kailash.maurya@microsoft.com}
}

\author{Pavan Mallapragada}
\affiliation{%
  \country{siva.mallapragada@microsoft.com}
}

\author{Yifan Shi}
\affiliation{%
  \country{yifanshi@microsoft.com}
}

\renewcommand{\shortauthors}{Jia et al.}



\settopmatter{printacmref=false}

\maketitle

\section{Introduction}
Importance sampling (IS) based policy estimators have been effective for counterfactual inference in complex systems~\cite{bottou2013counterfactual, mcbook2013is}.
Large-scale recommender systems, like ads auction systems~\cite{bottou2013counterfactual,gorurautomated}, utilize IS to evaluate parameters off-policy, aiming to achieve better key performance indicator (KPI) like revenue, clicks, and ad relevance. To identify a counterfactual distribution $p$ that better optimizes KPI values $f(x)$, Gorur et al.~\cite{gorurautomated} employ Gaussian proposal distribution $q$ and estimate $E[f(X)]$ under multiple alternative (\textit{counterfactual}) distributions $p$. Specifically, they sample tunable parameters $x \sim q$, observe KPI values $f(x)$, and estimate the KPI values as if the parameters were sampled from $p$ \textit{via}
\vspace{-0.05in}
\begin{equation}
\label{eqn:is}
E_{p}[f(X)] = E_{q}\big[f(X)\frac{p(X)}{q(X)}\big]\approx \frac{1}{N}\sum_{i=1}^{N}f(x_i)\frac{p(x_i)}{q(x_i)}.
\end{equation}
\vspace{-0.05in}

Randomizing $q$ such that it overlaps with $p$ as much as possible is crucial for obtaining accurate "what if" results, as the variance of the estimates increases when $p$ differs substantially from $q$~\cite{mcbook2013is}. However, simply broadening proposal coverage will increase randomization cost~\cite{gorurautomated}. 
The limitation is partially addressed by iteratively updating the proposal policy ~\cite{elvira2021advances,elvira2019generalized}.
Cornuet et al. ~\cite{cornuet2011adaptivemultipleimportancesampling} proposes the Adaptive Multiple Importance Sampling method that updates the importance weights for each sample across all past and present iterations. 
However, due to limited exploration in the proposal distribution at each iteration, the existing framework fails to overcome the issue of entrapment in local optima when doing iterative optimization to identify near-optimal parameters.

We propose a novel scheme using mixture distribution as the proposal distribution and dynamically adjusting both the mixture distributions parameters and their mixing rates at each iteration, enhancing search diversity and efficiency.
Through extensive simulations with varying noise levels, we demonstrate that the proposed policies converge better and faster to global optima with less regret. Furthermore, we validate that our proposed approaches are capable of identifying more effective tuning points in real-world ads auction tuning tasks on a major search engine platform.

\vspace{-0.05in}
\section{Adaptive Mixture IS}
\subsection{Gaussian Mixture Proposal Distribution}
To address the challenge of finding better local optima without simply increasing $\sigma_{q}$, we propose two policy variants under Gaussian Mixture proposal distribution framework (GMM). Both variants aim to effectively expand the search range: the multi-variance policy extends the search from the current proposal mean with controlled randomization costs; the multi-modal policy initiates the search from multiple targeted and restricted regions. 

Consider the GMM proposal distribution: $q(x) = \sum_{k=1}^K \pi_k\mathcal{N}(x\mid \mu_k, \sigma_k)$, where $K$ is the number of mixture components (peak and mixing rate pair) of this distribution.
$\pi \in R^K$ is a vector representing weight for each component such that $\sum_k \pi_k = 1$. 


\vspace{-0.05in}
\subsection{Multi-Variance Policy}
Inspired by $\epsilon$-greedy policy from reinforcement learning, where agent is allowed to take random actions to keep exploring for a small percentage of the trials \cite{sutton2018reinforcement}, we impose two Gaussian distributions where mean of both distributions is the same, but the second distribution has fatter tails: 
$p(z) = \pi_1 \mathcal{N}(\mu,\sigma_1) + \pi_2 \mathcal{N}(\mu,\sigma_2)$. 
Here, $K=2$, $\sigma_2 > \sigma_1$, and $\pi_2 < \pi_1$.  Figure~\ref{multivar} in Appendix \ref{app:multivar_appendix} illustrates how $\pi = [\pi_1, \pi_2]$ controls the randomization cost of the policy. 

\vspace{-0.05in}
\subsection{Multi-Modal Policy}
A possible alternative policy is to introduce $K$ different exploration peaks with the same variance. This strategy is advantageous when the KPI versus parameter landscape is multi-modal, whereas the current proposal mean is trapped at a local optimum, as exploring different peaks may uncover a better local optimum. It is particularly effective when the selection of initial peaks is informative. Appendix \ref{app:initial_peak} discusses the initial peaks selection in absence of priors.
\vspace{-0.05in}
\subsubsection{Score-based Greedy Update (GU)}
Algorithm \ref{alg:greedyPeak} employs a greedy best-second selection strategy based on rank scores. Starting with $K$ initial peak and mixing rate pairs, the algorithm iteratively selects counterfactual points. It first retains the point with the highest estimated value. Subsequently, it chooses the second counterfactual point that is at least $L_1$ distance $d$ away from the first point. 
While the mixing rates $\pi$ can be fixed, we propose a proportional allocation of 
$\pi$ according to rank scores to better leverage the scores and enhance the policy update. 
This approach can be further optimized by considering the top-$k$ candidates while maintaining at minimum distance $d$ to each other \cite{hong2018div-driven}.

Let $C$ be a collection of counterfactual points, $R$ be the corresponding estimated counterfactual KPI scores. 
Algorithm~\ref{alg:greedyPeak} then returns $O$ and $P$ i.e., the peaks and mixing rates respectively of the GMM proposal distribution.

\vspace{-0.08in}
\begin{algorithm}[H]
\caption{Score-based Greedy Update (GU)}
\label{alg:greedyPeak}
\begin{algorithmic}[H]
\State {\bfseries Input:} $C$, $R$, $K$, $\pi$, $d$
\State Sort $C$ based on the corresponding $R$ in descending order
\State Add the optimal candidate $c^* = C[0]$ to $O$, $r^*$ to $\pi$
\While {size of $O$ < $K$}
\For{each candidate ($c$, $r$) in ($C$, $R$)}
\If{$||c^*-c||_1 > d$ and $||o-c||_1 > d$ $\forall  o \in O$}
\State Add $c$ to $O$, $r$ to $\pi$
\EndIf
\EndFor
\EndWhile
\State Normalize $\pi$ such that $\sum \pi_i = 1$ \\
\Return $O$, $\pi$
\end{algorithmic}
\end{algorithm}
\vspace{-0.2in}

\subsubsection{Rank-based Peak Cluster Update (PCU)}

Instead of allocating the mixing rate $\pi$ at each iteration according to the actual rank score values, Algorithm \ref{alg:mixUpdate} uses the order of rank score to update $\pi$, which turns out to be less sensitive to noise in offline  simulation (see Table~  \ref{sim_result_main}). Specifically, the peak with the highest rank gets boosted in weight by $\delta$, and the entire $\pi$ will get normalized to sum 1. The multi-modal candidates at the next iteration will get nominated within each peak cluster. 
\vspace{-0.08in}
\begin{algorithm}[H]
\caption{Rank-based Peak Cluster Update (PCU)}
\label{alg:mixUpdate}
\begin{algorithmic}[H]
    \State {\bfseries Input:} $C$, $R$, $K$, $\pi$, $\delta$
    \For{each component $k$ in $(1, K)$}
        \State Subset $C_k$ as all counterfactual points generated by peak cluster $k$, $R_k$ as their corresponding counterfactual estimates
        \State Sort $C_k$ based on $R_k$ in descending order
        \State Add the optimal candidate ${C_k}[0]$ to $O$, estimate ${R_k}[0]$ to $P$
    \EndFor
    \State $\pi_k = \pi_k + \delta $, where $k = \arg\max P_i$
    \State Normalize $\pi$ such that $\sum \pi_i = 1$ \\
    \Return $O$, $\pi$
\end{algorithmic}
\end{algorithm}

\vspace{-0.2in}
\section{Experiments and Results}
We replicate the ad-tuning process through offline simulations, comparing Gaussian Importance Sampling (GIS) with Multi-variance update (MVU), and Multi-modal algorithms. Parameter samples $x$ are drawn from either a normal or mixture normal distribution, depending on the algorithm, and KPI values are obtained from a predefined generating function to emulate online ads experiments.



\subsection{Offline Simulation}

Our experiments demonstrate successful convergence for all four algorithms in absence of noise ($\gamma$=0) as shown in Table~\ref{sim_result_main}. For multi-modal algorithms specifically, we determine the convergence error using the peak with the highest mixing rate at the last iteration. While AMIS algorithms don't show significant improvement over GIS in terms of average regret and convergence error, they exhibit faster convergence speed. This can be attributed to their inherent increased randomization and exploration of a larger search space, which, while advantageous in noisy environments, may not provide a significant benefit in this simple scenario.

As the noise ratio increases to $\gamma$=10, GIS converges only 65\% of the times. All AMIS algorithms exhibit higher convergence rate, lower average regret and convergence error, and faster convergence across 100 runs. When $\gamma$=100, the convergence of all algorithms is compromised, yet the AMIS algorithms still notably outperforms GIS. The results show that AMIS approaches demonstrate superior robustness in noisy environments, 
which is particularly helpful as online ads auction environment constantly fluctuate with market conditions, user traffic, advertiser budget, etc.

While Table~\ref{sim_result_main} suggests improved convergence for multi-modal algorithms compared to multi-variance approach, real-world online experiments highlight their high sensitivity to the selection of initial peaks. The practical consideration of initial peak, effective sample size, and detailed experiment set-up are deferred to Appendix B.
\vspace{-0.1in}
\begin{table}[H]
  \large
  \resizebox{\columnwidth}{!}{\begin{tabular}{c|ccccccc}
     \toprule
  	  & Algo & Regret & \multicolumn{5}{c}{Convergence} \\
    \cmidrule(l){3-3}\cmidrule(l){4-8}
    & & MEAN & MAE & MSE & VAR & FCI & PRC \\
    \cmidrule(r){1-1}\cmidrule(l){2-8}
  	$\gamma$=0 & GIS & 11.074 & 0.094 & \textbf{0.019} & \textbf{0.019} & 5 & 100\% \\ 
  	$ESS$=0 & MVU & 10.289 & 0.200 & 0.040 & 0.040 & \textbf{3}  & 100\%\\ 
     & GU & 12.444 & 0.112 & 0.024 & 0.024 & \textbf{3}  & 100\%\\ 
     & PCU & \textbf{9.944} & \textbf{0.084} & 0.020 & 0.020 & 4  & 100\%\\  
    [0.1cm]\hline 
  	$\gamma$=10 & GIS & 17.133 & 2.488 & 17.851 & 11.898 & 5.871  & 65\%\\ 
  	$ESS$=0 & MVU & 12.334 & 1.284 & 13.224 & 11.961 & 3.863  & 87\% \\ 
     & GU & \textbf{11.067} & \textbf{0.236} & \textbf{1.470} & \textbf{1.453} & \textbf{3.323}  & \textbf{100\%} \\ 
     & PCU & 12.139 & 0.364 & 1.926 & 1.859 & 4.323  & \textbf{100\%} \\ 
    [0.1cm]\hline 
    $\gamma$=100 & GIS & 21.088& 4.310 & 28.847 & 11.591 & 7.067 & 15\% \\ 
  	$ESS$=0 & MVU & 16.812 & 3.856 & 31.742 & 23.052 & \textbf{5.102}  & 49\%\\ 
     & GU & \textbf{16.685} & 3.136 & 21.903 & 15.039 & 5.614  & \textbf{57\%}\\ 
     & PCU & 17.960 & \textbf{2.208} & \textbf{15.294} & \textbf{12.294} & 6.463  & 54\%\\
    [0.1cm]\bottomrule
  \end{tabular}}
  \caption{Regret and Convergence Metrics \normalfont for sample size $N = 100$, iterations $T = 10$, and number of repeated runs $R = 100$. \textbf{Mean regret} = $\frac{1}{RT}\sum_{r,t}{f(x=ground\ truth) - f(x_{r,t})}$. \textbf{FCI} (First Convergence Iteration): the first iteration where the last peak value is $\pm 0.02$ of the ground truth value 10. \textbf{PRC} (Percentage of Runs Converged): the number of converged runs over all repeated runs.
  }
  \label{sim_result_main}
\end{table}

\vspace{-0.4in}
\subsection{Online Experiments}
Beyond successful offline simulations, our AMIS approaches effectively identify optimal tuning points in an online A/B experiment on a major search engine.
Table~\ref{online_perf} presents an example mainstreamed tuning point found by multi-variance algorithm compared to the control group. Our online experiments demonstrate that tuning points generated by the AMIS algorithms are more likely to be adopted as mainstream configurations. 
We anticipate that the adoption of AMIS will enhance the accuracy and reliability of decision-making in the context of importance sampling off-policy estimators.

\vspace{-0.1in}
\begin{table}[H]
\large
  \resizebox{\columnwidth}{!}{
  \begin{tabular}{c|c|c|c|c}
    \toprule
    & Revenue & Clicks & AdSpace & AdRelevance \\ \hline
  	Snapshot & 0.88\% & -0.42\% & -2.29\% & 1.10\% \\ \hline
    Holdout & 1.16\% & -0.40\% & -2.26\% & 1.35\%\\ 
    \bottomrule
  \end{tabular}
\  }
  \caption{Performance of an example tuning point.}
  \label{online_perf}
  \vskip -0.2in  
\end{table}
\vspace{-0.1in}


\begin{acks}
The authors are grateful to many colleagues for helpful discussions on solving the ads auction problem, in particular to Adith Swaminathan, for untiring reviews and guidance, and to Sendill Arunachalam and Ming Li for engineering support. 
\end{acks}
\bibliographystyle{acm}  
\bibliography{main}

\newpage
\appendix
\section{Mixture Importance Sampling Illustration}
\subsection{Multi-variance policy}
\label{app:multivar_appendix}
\begin{figure}[H]
\centering
{\includegraphics[scale=0.55]{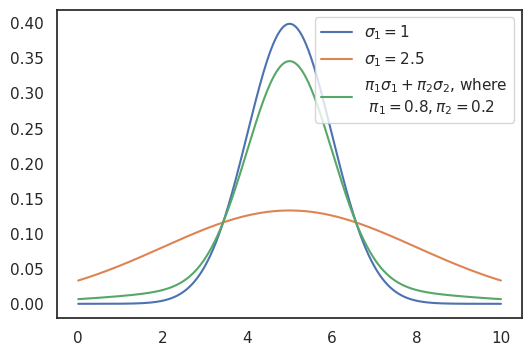}}
\caption{Multi-variance Policy: Exploit \& Exploration Variances Centered at the Same Mean}
\label{multivar}
\end{figure}
Indiscriminately increasing the proposal scale parameter, $\sigma_q$, will result in a more diffuse parameter distribution, potentially leading to very unexpected performances. As illustrated in  Figure \ref{multivar}, even with $\sigma_q=2.5$, the distribution exhibits a more pronounced short and fat-tailed shape than $\sigma_q=1$. 

In contrast, our proposed multi-variance policy, with $\pi = [0.8, 0.2]$ maintains the parameters at their current reasonable values most of the time while allowing for occasional exploration through a fatter tail. The amount of exploration and thus randomization cost is under control over the $\pi$ vector.

\subsection{Multi-modal policy}
\label{app:multimodal_appendix}
\begin{figure}[H]
\centering{\includegraphics[scale=0.7]{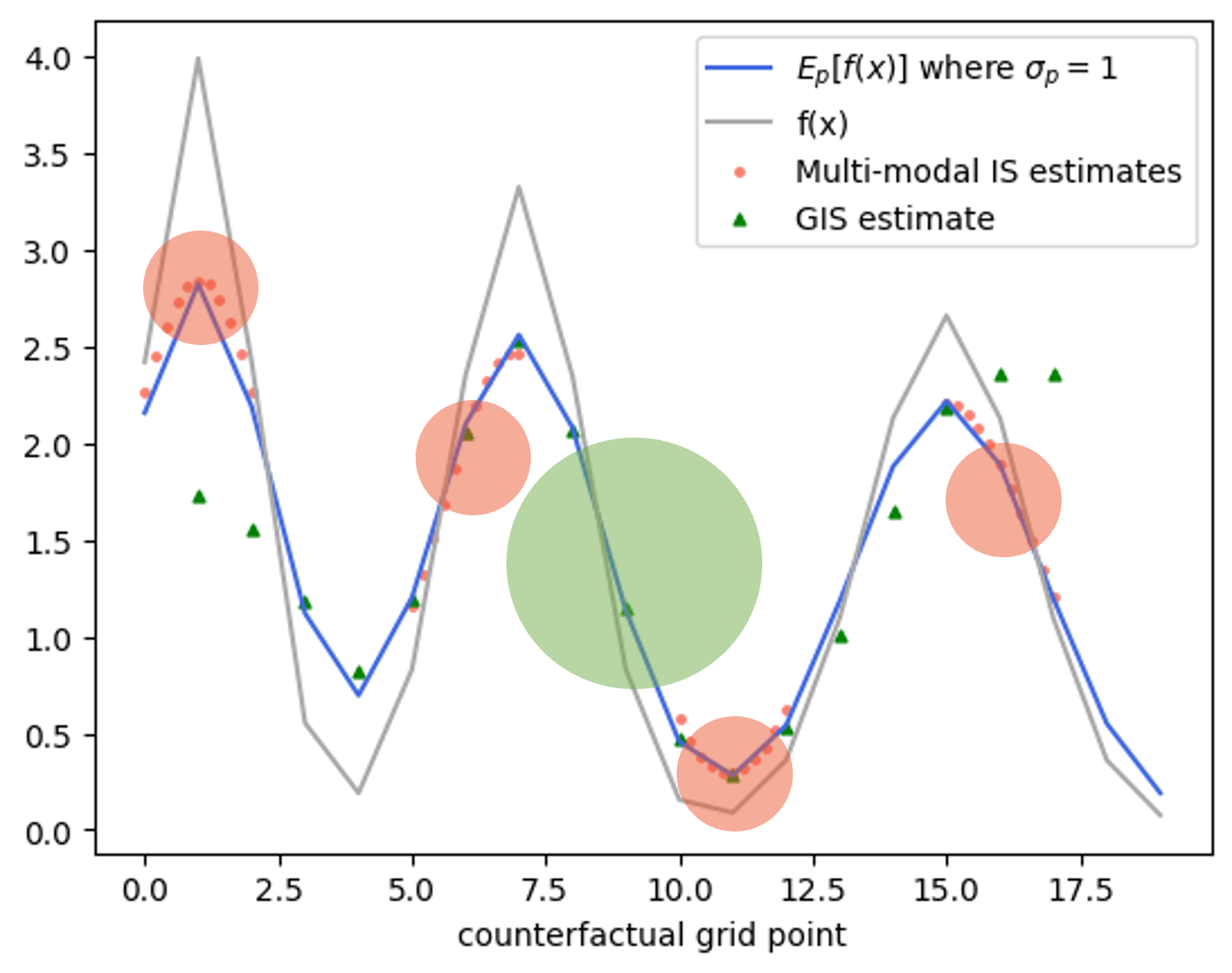}}
\caption{Multi-modal Policy: Explore with a large peak vs many small peaks. }
\label{multipeak}
\end{figure}

In online ads auction tuning setting, we aim to find $\arg\max f(x)$, or the local optimum of $f(x)$ in a reasonable area, where $f(x)$ is the KPI over tunable parameter distribution. As depicted in an importance sampling simulation Figure \ref{multipeak}, $f(x)$ and $E_p[f(X)]$ exhibit similar local and global maxima when counterfactual $\sigma_p = 1$. 

Evaluating nearby counterfactual points using the small red circles centered at 1, 6, 11, and 16, with $\sigma_q = \sigma_p = 1$, provide accurate estimates of $E_p[f(x)]$: the red dots closely align with the blue line, helping us get a good understanding of the landscape at those local regions. 

In contrast, evaluating the importance sampling estimator from the larger green circle centered at 9 even with increased proposal scale $\sigma_q = 2$ ($\sigma_q$ remains at 1), can lead to substantial error estimates when the estimated region deviates from the proposal distribution. For instance, when the counterfactual grid point falls outside the range of 1 to 16, the green triangles deviate from the blue line. This occurs despite the larger proposal sampling range, which might be expected to increase randomization costs.

This simple simulation illustrate that searching from multiple small peaks (the multi-modal policy) can lead to better and more rapid convergence to the global optimum at 1, compared to searching from one large peak.

%

\section{Case Study: Parameter Tuning in Ads Auction}

Major search engines employ Generalized Second Price (GSP) auctions to allocate ad placements on sponsored search pages \cite{varian2007position}. In this mechanism, ads are ranked based on a score function, with the highest-scoring ad occupying the top position. The ad's price is determined by the minimum bid necessary to retain its position. The rank score for ad i, denoted as $S_i(b_i)$, can be approximated as:
$$S_i(b_i) = b_iQ_i = b_i \sum_j \lambda_{ij}q_{ij}$$
Here,  $b_i$ is the bid, $Q_i$ is the weighted quality score influenced by factors such as probability of click and ad relevance, $\lambda_{ij}$ are auxiliary weight parameters for each quality score $q_{ij}$.

Auxiliary weight parameters significantly influence ad ranking and, consequently, key performance indicators (KPIs) such as revenue per thousand impressions (RPM), click-through rate (CTR), and ad relevance. To optimize overall performance, these parameters require continuous adjustment in response to market dynamics and model updates. The process of identifying optimal values for $\lambda_{ij}$ that yield substantial improvements in overall KPIs while other tuning environment variables are maintained constant is referred to as \textit{tuning}. Typically, different devices and markets necessitate distinct tunable parameters and weights, rendering parameter tuning a complex challenge due to both the exponentially increasing number of potential tunings and the continuous nature of the parameters. To mitigate these challenges, Adaptive Importance Sampling with a mixture distribution is employed to expand the search space and expedite the convergence to superior tuning points.

\begin{figure}
\centering
{\includegraphics[scale=0.3]{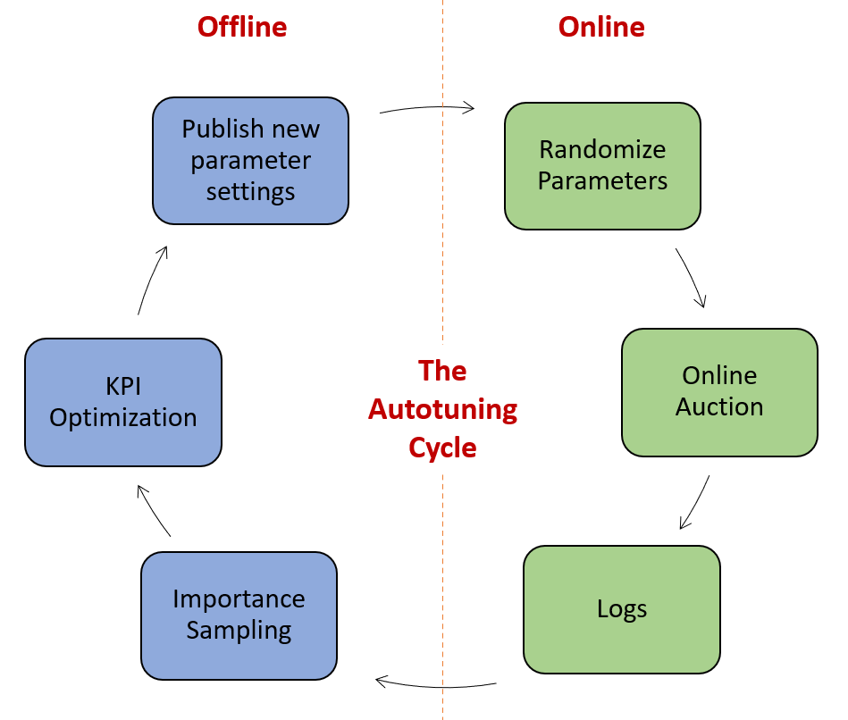}}
\caption{Automated tuning cycle}
\label{autotune}
\end{figure}

\subsection{Simulating Ads Auction Tuning}
\label{app:ad_tuning}

We replicate the automated tuning cycle delineated in Figure \ref{autotune} through offline simulations. Specifically, we have implemented the following algorithms: Importance Sampling with Gaussian proposal distribution (GIS), Score-based Greedy Update (GU) as described in Algorithm \ref{alg:greedyPeak},Rank-based Peak Cluster Update (PCU) detailed in Algorithm \ref{alg:mixUpdate}, and the Multi-Variance Update (MVU) policy as described in section 2.2. 

To avoid an explosion in the number of mixture parameter combinations, we use simple Gaussian distributions as counterfactual distributions. For each $\mu_k$, we create counterfactual grids with means at multiples of $\sigma_k$ around $\mu_k$. For example: $\{\mu_k - \sigma_k, \mu_k-0.9 \sigma_k, ..., \mu_k, \mu_k + 0.1 \sigma_k, ..., \mu_k + \sigma_k\}$. Each proposal candidate is evaluated using Eqn \ref{eqn:is} and are ranked based on their estimates. 

Unlike real online experiments, where the KPI over parameters landscape $f(x)$ is observed through A/B experiment, the $f(x)$ in our study is simulated. We model it using a normal distribution with added noise:
$$f_{KPI}(x) = \frac{1}{\sigma\sqrt{2\pi}} 
  \exp\left( -\frac{1}{2}\left(\frac{x-\mu}{\sigma}\right)^{\!2}\,\right) \times 100 + N(0, \gamma \sigma) , $$

where $\gamma$ is the noise ratio. We magnifying the density with a constant of 100 to show enough distinction between density values with respect to Importance Sampling weights values. Note that when the true landscape of KPI versus parameter space is "multi-modal" with multiple peaks or spikes, the best strategy $E_p[f(x)]$ will converge to somewhere in between the peaks, because Importance Sampling needs to maintain exploration through the variance of a Normal or mixutures of Normal distributions. 

Unlike density estimation tasks, our goal is not to identify all the peaks, but to find better local optima. We focus on scenarios where the KPI vs. parameter space exhibits a single peak.
Table \ref{hyperparam} shows the hyper-parameters set for this simulation.

\begin{table}[H]
  \resizebox{\columnwidth}{!}{
  \begin{tabular}{c|c|c|c|c}
    \toprule
    & VIS & PU & MRU & MVU \\ \hline
  	Initial Peak(s) & 5 & \multicolumn{2}{c|}{ [3, 5, 7]} & 5 \\ \hline
    Initial Mixing Rates & 1 & \multicolumn{2}{c|}{ [0.2, 0.6, 0.2]} & [0.8, 0.2]\\ \hline
    Proposal $\sigma$ & 1 &  \multicolumn{2}{c|}{ 1} & [1, 3]\\ \hline
    Counterfactual $\sigma$ & 1 & \multicolumn{2}{c|}{ 1} & 2 \\ \hline 
    Peak Distance Coefficient & NA & 1.5 & NA & NA \\ \hline
    Ground truth mean & \multicolumn{4}{c}{ 10} \\ \hline
    Ground truth $\sigma$ & \multicolumn{4}{c}{ 1}  \\ \hline
    Counterfactual Grid Size & \multicolumn{4}{c}{ 11} \\
    \bottomrule
  \end{tabular}
\  }
  \caption{Hyper-parameters for offline simulation}
  \label{hyperparam}
\end{table}

\subsection{Hyperparameter Considerations}

\subsubsection{Initial Peaks Selection}
\label{app:initial_peak}
While simulations in Table~\ref{sim_result} suggest lower regret and improved convergence for multi-peak algorithms compared to the multi-variance approach, real-world online experiments highlight their high sensitivity to the selection of initial peaks. Without prior knowledge, the initial minor peaks can be chosen at the orthogonal edges of the high-dimensional parameter space, as inspired by studies on orthogonal experimental design \cite{Xu2013COMBININGTA}. In online settings, the number of tunable parameters is around 5. To comprehensively explore all potential directions towards the global optimum relative to the current point, one would ideally require $2^5=32$ initial peaks to ensure that at least one minor peak is closer to the ground truth than the initial major peak. However, with this number of initial peaks, each minor peak would have a correspondingly less than 0.1 mixing rate. In practice, computational constraints often necessitate selecting a smaller number of peaks, typically from 4 to 9. If, by chance, none of these minor peaks are closer to the global optimum than the initial major peak, the multi-modal algorithm can actually perform worse than the multi-variance approach in terms of regret and convergence speed.

\subsubsection{Effective sample size and confidence interval bound}
\label{app:ess}
\begin{figure}
\centering
{\includegraphics[scale=0.8]{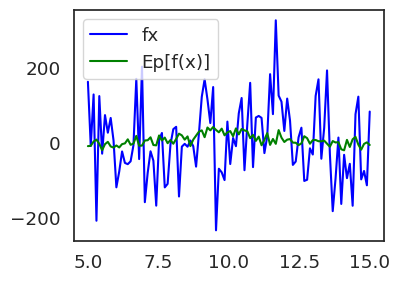}}
\caption{An Example Ground Truth Plot at Noise Ratio $\gamma$=100}
\label{fx_n100}
\end{figure}
Effective sample size (ESS) is a term used in survey statistics to describe the number of independent samples in a dataset that are equivalent to the original dataset in terms of statistical power. In the context of importance sampling, ESS is employed to evaluate the adequacy of the weight distribution \cite{mcbook2013is}. A larger ESS corresponds to increased precision in parameter estimation. The ESS in this context is calculated as:
$ESS = \frac{(\sum w_{i})^2}{\sum w_{i}^2}$. While increasing the ESS threshold eliminates the counterfactual points with higher weight variances, it concurrently limits the step size at each iteration and induces more pronounced fluctuations in the convergence rate and thus  higher regret. Empirical experiments demonstrate under high noise when $\gamma=100$ (see Plot \ref{fx_n100} and Table \ref{sim_result}) that incorporating an ESS constraint can diminish convergence variance. However, it does not guarantee to improve convergence speed or to mitigate the regret.  

In addition to ESS, assessing the variance of importance sampling estimates is crucial \cite{mcbook2013is}. Inspired by Upper Confidence Bound (UCB) methodology \cite{bubeck2012regretanalysisstochasticnonstochastic}, confidence intervals are integrated into counterfactual point selection to account for variance in online experiments. It is essential to recognize that variance estimation relies on the same weights used for mean estimation. Consequently, disproportionately skewed weights can distort both mean and variance estimates, potentially masking underlying issues \cite{mcbook2013is}.

\begin{table}
  \large
  \resizebox{\columnwidth}{!}{\begin{tabular}{c|ccccccc}
     \toprule
  	  & Algo & Regret & \multicolumn{5}{c}{Convergence} \\
    \cmidrule(l){3-3}\cmidrule(l){4-8}
    & & MEAN & MAE & MSE & VAR & FCI & PRC \\
    \cmidrule(r){1-1}\cmidrule(l){2-8}
    $\gamma$=100 & VIS & 21.088& 4.310 & 28.847 & 11.591 & 7.067 & 15\% \\ 
  	$ESS$=0 & MVU & 16.812 & 3.856 & 31.742 & 23.052 & \textbf{5.102}  & 49\%\\ 
     & PU & \textbf{16.685} & 3.136 & 21.903 & 15.039 & 5.614  & \textbf{57\%}\\ 
     & MRU & 17.960 & \textbf{2.208} & \textbf{15.294} & \textbf{12.294} & 6.463  & 54\%\\
    [0.1cm]\hline 
  	$\gamma$=100 & VIS & 21.861 & 4.266 & 26.7428 & \textbf{9.019} & 7.200  & 15\%\\ 
  	$ESS$=0.4 & MVU & 20.593 & 4.336 & 32.277 & 15.368 & 7.033  & 30\% \\ 
     & PU & \textbf{16.303} & 2.872 & 18.2592 & 13.188 & \textbf{5.404}  & \textbf{57\%} \\ 
     & MRU & 18.464 & \textbf{2.567} & \textbf{16.528} & 11.819 & 6.74  & 50\% \\ 
    [0.1cm]\bottomrule
  \end{tabular}}
  \caption{Regret and Convergence Metrics \normalfont for sample size $N = 100$, number of iteration $T = 10$, and number of repeated runs $R = 100$. \textbf{Mean regret} = $\frac{1}{RT}\sum_{r,t}{f(x=groud\ truth\ mean) - f(x_{r,t})}$. \textbf{FCI} (First Convergence Iteration): the first iteration where the last peak value is $\pm 0.02$ close to the ground truth value 10. \textbf{PRC} (Percentage of Runs Converged): the number of converged runs over all repeated runs.
  }
  \label{sim_result}
\end{table}



\newpage






\end{document}